\documentclass[11pt]{article}
\usepackage{acl2016}

\usepackage{times}
\usepackage[skip=0pt]{caption}

\usepackage[table]{xcolor}
\usepackage{color,graphicx}
\usepackage{latexsym,multirow,framed,amsmath,amsfonts,amssymb,url}
\usepackage{centernot}
\usepackage{booktabs}

\newcommand{\ignore}[1]{}
\usepackage{todonotes}

\aclfinalcopy 

\usepackage{array}
\newcolumntype{L}[1]{>{\raggedright\let\newline\\\arraybackslash\hspace{0pt}}m{#1}}
\newcolumntype{C}[1]{>{\centering\let\newline\\\arraybackslash\hspace{0pt}}m{#1}}
\newcolumntype{R}[1]{>{\raggedleft\let\newline\\\arraybackslash\hspace{0pt}}m{#1}}

\date{}

\newif\iflogvar
\logvartrue

\title{
\vspace*{-0.5in}
{\small \hfill \textit{To appear in ACL'16}}
\\
\vspace*{.25in} 
Two Discourse Driven Language Models for Semantics
}
 
\author{Haoruo Peng $\;$ \textnormal{and} $\;$ Dan Roth \\
	    University of Illinois, Urbana-Champaign \\
	    Urbana, IL, 61801 \\
	    {\tt \{hpeng7,danr\}@illinois.edu}}

\date{}

\begin{document}
\maketitle
\begin{abstract}
Natural language understanding often requires deep semantic knowledge. Expanding on previous proposals, we suggest that some important aspects of semantic knowledge can be modeled as a language model if done at an appropriate level of abstraction. 
We develop two distinct models that capture semantic frame chains and discourse information while abstracting over the specific mentions of predicates and entities. 
For each model, we investigate four implementations: a ``standard" N-gram language model and three discriminatively trained ``neural" language models that generate embeddings for semantic frames. The quality of the semantic language models (SemLM) is evaluated both intrinsically, using perplexity and a narrative cloze test and extrinsically -- we show that our SemLM helps improve performance on semantic natural language processing tasks such as co-reference resolution and discourse parsing. 
\end{abstract}

\section{Introduction} \label{sec:intro}
Natural language understanding often necessitates deep semantic knowledge. This knowledge needs to be captured at multiple levels, from words to phrases, to sentences, to larger units of discourse. At each level, capturing meaning frequently requires  context sensitive abstraction and  disambiguation, as shown in the following  example~\cite{winograd1972understanding}:

\vspace{0.05in}
\parbox{2.9in}{Ex.1 \textit{[Kevin] was \textbf{robbed} by [Robert]. [He] was \textbf{arrested} by the police.} \\ 
Ex.2 \textit{[Kevin] was \textbf{robbed} by [Robert]. [He] was \textbf{rescued} by the police.}} \vspace{0.01in} \\
\ignore{               
\vspace{0.05in}
\parbox{2.9in}{Ex.1 \textit{[Robert] \textbf{robbed} [Kevin], \underline{and} the police \textbf{arrested} [him].} \\ 
               Ex.2 \textit{[Robert] \textbf{robbed} [Kevin], \underline{and} the police \textbf{rescued} [him].}} \vspace{0.01in} \\}               
In both cases, one needs to resolve the pronoun ``he'' to either ``Robert'' or ``Kevin''. To make the correct decisions, one needs to know that the subject of ``rob'' is more likely than the object of ``rob" to be the object of ``arrest'' while the object of ``rob'' is more likely to be the object of ``rescue''. Thus, beyond understanding individual predicates (e.g., at the semantic role labeling level), there is a need to place them and their arguments in a global context. 

However, just modeling semantic frames is not sufficient; consider a variation of Ex.1: 

\vspace{0.05in}
\parbox{2.9in}{Ex.3 \textit{Kevin was \textbf{robbed} by Robert, \underline{but} the police mistakenly \textbf{arrested} him.}} \vspace{0.01in} \\
In this case, ``him'' should refer to ``Kevin'' as the discourse marker ``but'' reverses the meaning, illustrating that it is necessary to take discourse markers into account when modeling semantics.

In this paper we propose that these aspects of semantic knowledge can be modeled as a {\em Semantic Language Model} (SemLM).
Just like the ``standard" syntactic language models (LM), we define a basic vocabulary, a finite representation language, and a prediction task, which allows us to model the distribution over the occurrence of elements in the vocabulary as a function of their (well-defined) context. In difference from syntactic LMs, we represent natural language at a higher level of semantic abstraction, thus facilitating modeling deep semantic knowledge.  

We propose two distinct discourse driven language models to capture semantics. In our first semantic language model, the {\em Frame-Chain SemLM}, we model all semantic frames and discourse markers in the text. Each document is viewed as a single chain of semantic frames and discourse markers. Moreover, while the vocabulary of discourse markers is rather small, the number of different surface form semantic frames that could appear in the text is very large. To achieve a better level of abstraction, we disambiguate semantic frames and map them to their PropBank/FrameNet representation. 
Thus, in Ex.3, the resulting frame chain is ``rob.01 --- but --- arrest.01'' (``01'' indicates the predicate sense). 

Our second semantic language model is called {\em Entity-Centered SemLM}. Here, we model a sequence of semantic frames and discourse markers involved in a specific co-reference chain. For each co-reference chain in a document, we first extract semantic frames corresponding to each co-referent mention, disambiguate them as before, and then determine the discourse markers between these frames. Thus, each unique frame contains both the disambiguated predicate and the argument label of the mention. In Ex.3, the resulting sequence is ``rob.01\#obj --- but --- arrest.01\#obj'' (here ``obj'' indicates the argument label for ``Kevin'' and ``him'' respectively). While these two models capture somewhat different semantic knowledge, we argue later in the paper that both models can be induced at high quality, and that they are suitable for different NLP tasks. 

For both models of SemLM, we study four language model implementations: N-gram, skip-gram~\cite{MikolovYiZw13}, continuous bag-of-words~\cite{mikolov2013efficient} and log-bilinear language model~\cite{mnih2007three}. Each model defines its own prediction task. In total, we produce eight different SemLMs. Except for N-gram model, others yield embeddings for semantic frames as they are neural language models. 

In our empirical study, we evaluate both the quality of all SemLMs and their application to co-reference resolution and shallow discourse parsing tasks. Following the traditional evaluation standard of language models, we first use perplexity as our metric. We also follow the script learning literature~\cite{chambers2008unsupervised,chambers2009unsupervised,rudinger2015script} and evaluate on the narrative cloze test, i.e. randomly removing a token from a sequence and test the system's ability to recover it. 
We conduct both evaluations on two test sets: a hold-out dataset from the New York Times Corpus and gold sequence data (for frame-chain SemLMs, we use PropBank~\cite{KingsburyPa02}; for entity-centered SemLMs, we use Ontonotes~\cite{HMPRW06} ). By comparing the results on these test sets, we show that we do not incur noticeable degradation when building SemLMs using preprocessing tools. Moreover, we show that SemLMs improves the performance of co-reference resolution, as well as that of predicting the sense of discourse connectives for both explicit and implicit ones.

The main contributions of our work can be summarized as follows: 1) The design of two novel discourse driven Semantic Language models, building on text abstraction and neural embeddings; 2) The implementation of high quality SemLMs that are shown to improve state-of-the-art NLP systems.


\section{Related Work} \label{sec:rel}
Our work is related to script learning. Early works~\cite{schank2013scripts,mooney1985learning} tried to construct knowledge bases from documents to learn scripts. Recent work focused on utilizing statistical models to extract high-quality scripts from large amounts of data~\cite{ChambersJu08,bejan2008unsupervised,jans2012skip,pichotta2014statistical,granroth2015happens,pichotta2016learning}. Other works aimed at learning a collection of structured events~\cite{chambers2013event,cheung2013probabilistic,cheung2013probabilistic,balasubramanian2013generating,bamman2014unsupervised,nguyen2015generative}, and several works have employed neural embeddings~\cite{modi:ICLR2014,modi:CONLL2014,fermann:EACL2014,titov-khoddam:2015:NAACL-HLT}. \newcite{ferraro2016unified} presented a unified probabilistic model of syntactic and semantic frames while also demonstrating improved coherence.

In our work, the semantic sequences in the entity-centered SemLMs are similar to narrative schemas~\cite{chambers2009unsupervised}. However, we differ from them in the following aspects: 1) script learning does not generate a probabilistic model on semantic frames\footnote{Some works may utilize a certain probabilistic framework, but they mainly focus on generating high-quality frames by filtering.}; 2) script learning models semantic frame sequences incompletely as they do not consider discourse information; 3) works in script learning rarely show applications to real NLP tasks.

Some prior works have used scripts-related ideas to help improve NLP tasks~\cite{irwin2011narrative,RahmanNg11c,peng-khashabi-roth:2015:NAACL-HLT}. However, since they use explicit script schemas either as features or constraints, these works suffer from data sparsity problems. In our work, the SemLM abstract vocabulary ensures a good coverage of frame semantics.

\section{Two Models for SemLM} \label{sec:unit}
\begin{table}
\captionsetup{belowskip=0pt,aboveskip=0pt}
\caption{\textbf{Comparison of vocabularies between frame-chain (FC) and entity-centered (EC) SemLMs.} ``F-Sen'' stands for frames with predicate sense information while ``F-Arg'' stands for frames with argument role label information; ``Conn'' means discourse marker and ``Per'' means period. ``Seq/Doc'' represents the number of sequence per document.}
\label{tab:naive-coref}
\centering
{
\begin{tabular}{l|c|c|c|c|c}
\toprule
       & F-Sen & F-Arg & Conn & Per & Seq/Doc \\ \hline 
FC  & YES   & NO    & YES  & YES	& Single      \\ \hline
EC  & YES   & YES   & YES  & NO  & Multiple    \\ \bottomrule
\end{tabular}
}
\vspace{-0.2in}
\end{table}

In this section, we describe how we capture sequential semantic information consisted of semantic frames and discourse markers as semantic units (i.e. the vocabulary).

\subsection{Semantic Frames and Discourse Markers}
\noindent {\bf Semantic Frames} A semantic frame is composed of a predicate and its corresponding argument participants. Here we require the predicate to be disambiguated to a specific sense, and we need a certain level of abstraction of arguments so that we can assign abstract labels. The design of PropBank frames~\cite{KingsburyPa02} and FrameNet frames~\cite{baker1998berkeley} perfectly fits our needs. They both have a limited set of frames (in the scale of thousands) and each frame can be uniquely represented by its predicate sense. These frames provide a good level of generalization as each frame can be instantiated into various surface forms in natural texts. We use these frames as part of our vocabulary for SemLMs. Formally, we use the notation $\text{f}$ to represent a frame. Also, we denote $\text{fa} \triangleq \text{f}\#\text{Arg}$ when referring to an argument role label (Arg) inside a frame (f).

\noindent {\bf Discourse Markers}
We use discourse markers (connectives) to model discourse relationships between frames. There is only a limited number of unique discourse markers, such as \textit{and, but, however}, etc. We get the full list from the Penn Discourse Treebank~\cite{PDLMRJW08} and  include them as part of our vocabulary for SemLMs. Formally, we use \textit{dis} to denote the discourse marker. Note that discourse relationships can exist without an explicit discourse marker, which is also a challenge for discourse parsing. Since we cannot reliably identify implicit discourse relationships, we only consider explicit ones here. More importantly, discourse markers are associated with arguments~\cite{WellnerPu07} in text (usually two sentences/clauses, sometimes one). We only add a discourse marker in the semantic sequence when its corresponding arguments contain semantic frames which belong to the same semantic sequence. We call them \textit{frame-related discourse markers}. Details on generating semantic frames and discourse markers to form semantic sequences are discussed in Sec.~\ref{sec:preprocess}.

\subsection{Frame-Chain SemLM}
For frame-chain SemLM, we model all semantic frames and discourse markers in a document. We form the semantic sequence by first including all semantic frames in the order they appear in the text: $[\text{f}_1, \text{f}_2, \text{f}_3, \ldots]$. Then we add \textit{frame-related discourse markers} into the sequence by placing them in their order of appearance. Thus we get a sequence like $[\text{f}_1, \text{dis}_1, \text{f}_2, \text{f}_3, \text{dis}_2, \ldots]$. Note that discourse markers do not necessarily exist between all semantic frames. Additionally, we treat the \textit{period} symbol as a special discourse marker, denoted by ``o''. As some sentences contain more than one semantic frame (situations like clauses), we get the final semantic sequence like this:
\[[\text{f}_1, \text{dis}_1, \text{f}_2, \text{o}, \text{f}_3, \text{o}, \text{dis}_2, \ldots, \text{o}]\]

\subsection{Entity-Centered SemLM}
We generate semantic sequences according to co-reference chains  for entity-centered SemLM. From co-reference resolution, we can get a sequence like $[\text{m}_1, \text{m}_2, \text{m}_3, \ldots]$, where mentions appear in the order they occur in the text. Each mention can be matched to an argument inside a semantic frame. Thus, we replace each mention with its argument label inside a semantic frame, and get $[\text{fa}_1, \text{fa}_2, \text{fa}_3, \ldots]$. We then add discourse markers exactly in they way we do for frame-chain SemLM, and get the following sequence: 
\[[\text{fa}_1, \text{dis}_1, \text{fa}_2, \text{fa}_3, \text{dis}_2, \ldots]\]

The comparison of vocabularies between frame-chain and entity-centered SemLMs is summarized in Table~\ref{tab:naive-coref}.

\section{Implementations of SemLM} \label{sec:slm}
In this work, we experiment with four language model implementations: N-gram (NG), Skip-Gram (SG), Continuous Bag-of-Words (CBOW) and Log-bilinear (LB) language model. For ease of explanation, we assume that a semantic unit sequence is $s = [w_1,w_2,w_3, \ldots, w_k]$.

\subsection{N-gram Model}
For an n-gram model, we predict each token based on its $n-1$ previous tokens, i.e. we directly model the following conditional probability (in practice, we choose $n=3$, Tri-gram (TRI) ):
\[ p(w_{t+2}|w_t, w_{t+1}). \]
Then, the probability of the sequence is
\[ p(s) = p(w_1)p(w_2|w_1)\prod_{t=1}^{k-2} p(w_{t+2}|w_t, w_{t+1}). \]

To compute $p(w_2|w_1)$ and $p(w_1)$, we need to back off from Tri-gram to Bi-gram and Uni-gram.

\subsection{Skip-Gram Model}
The SG model was proposed in~\newcite{MikolovYiZw13}. It uses a token to predict its context, i.e. we model the following conditional probability:
\[ p(c\in c(w_t)|w_t, \theta). \]
Here, $c(w_t)$ is the context for $w_t$ and $\theta$ denotes the learned parameters which include neural network states and embeddings. Then the probability of the sequence is computed as
\[ \prod_{t=1}^k \prod_{c\in c(w_t)} p(c|w_t, \theta). \]

\subsection{Continuous Bag-of-Words Model}
In contrast to skip-gram, CBOW~\cite{mikolov2013efficient} uses context to predict each token, i.e. we model the following conditional probability:
\[ p(w_t|c(w_t), \theta). \]
In this case, the probability of the sequence is
\[ \prod_{t=1}^k p(w_t|c(w_t),\theta). \]

\subsection{Log-bilinear Model}
LB was introduced in~\newcite{mnih2007three}. Similar to CBOW, it also uses context to predict each token. However, LB associates a token with three components instead of just one vector: a target vector v(w), a context vector v'(w) and a bias b(w). So, the conditional probability becomes:
\[ p(w_t|c(w_t)) = \frac{\exp(v(w_t)^\intercal u(c(w_t))+b(w_t))}{\sum_{w\in \mathcal{V}}\exp(v(w)^\intercal u(c(w_t))+b(w))}. \]
Here, $\mathcal{V}$ denotes the vocabulary and we define $u(c(w_t))=\sum_{c_i\in c(w_t)} q_i \odot v'(c_i)$. Note that $\odot$ represents element-wise multiplication and $q_i$ is a vector that depends only on the position of a token in the context, which is a also a model parameter. 

So, the overall sequence probability is
\[ \prod_{t=1}^k p(w_t|c(w_t)). \]

\section{Building SemLMs from Scratch} \label{sec:preprocess}
In this section, we explain how we build SemLMs from un-annotated plain text.

\subsection{Dataset and Preprocessing}
\noindent {\bf Dataset} We use the New York Times Corpus\footnote{https://catalog.ldc.upenn.edu/LDC2008T19} (from year 1987 to 2007) for training. It contains a bit more than 1.8M documents in total.

\noindent {\bf Preprocessing} We pre-process all documents with semantic role labeling~\cite{PRYZ04} and part-of-speech tagger~\cite{RothZe98}. We also implement the explicit discourse connective identification module in shallow discourse parsing~\cite{SPKSR15}. Additionally, we utilize within document entity co-reference~\cite{PengChRo15} to produce co-reference chains. To obtain all annotations, we employ the Illinois NLP tools\footnote{http://cogcomp.cs.illinois.edu/page/software/}.

\subsection{Semantic Unit Generation}
\noindent {\bf FrameNet Mapping} We first directly derive semantic frames from semantic role labeling annotations. As the Illinois SRL package is built upon PropBank frames, we do a mapping to FrameNet frames via VerbNet senses~\cite{schuler2005verbnet}, thus achieving a higher level of abstraction. The mapping file\footnote{http://verbs.colorado.edu/verb-index/fn/vn-fn.xml} defines deterministic mappings. However, the mapping is not complete and there are remaining PropBank frames. Thus, the generated vocabulary for SemLMs contains both PropBank and FrameNet frames. For example, ``place'' and ``put'' with the VerbNet sense id ``9.1-2'' are converted to the same FrameNet frame ``Placing''.

\noindent {\bf Augmenting to Verb Phrases} We apply three heuristic modifications to augment semantic frames defined in Sec. 3.1: 1) if a preposition immediately follows a predicate, we append the preposition to the predicate e.g. ``take over''; 2) if we encounter the semantic role label AM-PRD which indicates a secondary predicate, we also append this secondary predicate to the main predicate e.g. ``be happy''; 3) if we see the semantic role label AM-NEG which indicates negation, we append ``not'' to the predicate e.g. ``not like''. These three augmentations can co-exist and they allow us to model more fine-grained semantic frames. 

\noindent {\bf Verb Compounds} We have observed that if two predicates appear very close to each other, e.g. ``eat and drink'', ``decide to buy'', they actually represent a unified semantic meaning. Thus, we construct compound verbs to connect them together. We apply the rule that if the gap between two predicates is less than two tokens, we treat them as a unified semantic frame defined by the conjunction of the two (augmented) semantic frames, e.g. ``eat.01-drink.01'' and ``decide.01-buy.01''.

\noindent {\bf Argument Labels for Co-referent Mentions} To get the argument role label information for co-referent mentions, we need to match each mention to its corresponding semantic role labeling argument. If a mention head is inside an argument, we regard it as a match. We do not consider singleton mentions.

\noindent {\bf Vocabulary Construction} After generating all semantic units for (augmented and compounded) semantic frames and discourse markers, we merge them together as a tentative vocabulary. In order to generate a sensible SemLM, we filter out rare tokens which appear less than 20 times in the data. We  add the Unknown token (UNK) and End-of-Sequence  token (EOS) to the eventual vocabulary.

\begin{table}
\captionsetup{belowskip=0pt,aboveskip=0pt}
\footnotesize
\caption{\textbf{Statistics on SemLM vocabularies and sequences.} ``F-s'' stands for single frame while ``F-c'' stands for compound frame; ``Conn'' means discourse marker. ``\#seq'' is the number of sequences, and ``\#token'' is the total number of tokens (semantic units). We also compute the average token in a sequence i.e. ``\#t/s''. We compare frame-chain (FC) and entity-centered (EC) SemLMs to the usual syntactic language model setting i.e. ``LM''.}
\label{tab:vocab}
\centering
{
\begin{tabular}{l|c|c|c|c|c|c}
\toprule
       & \multicolumn{3}{c|}{Vocabulary Size} & \multicolumn{3}{c}{Sequence Size} \\
       & F-s   & F-c   & Conn & \#seq & \#token & \#t/s            \\ \hline 
FC  & 14857 & 7269  & 44   & 1.2M  & 25.4M   & 21               \\
EC  & 8758  & 2896  & 44   & 3.4M  & 18.6M   & 5                \\ \hline
LM  & \multicolumn{3}{c|}{$\sim$20k}  & $\sim$3M  & $\sim$38M  & 10-15  \\
\bottomrule
\end{tabular}
}
\vspace{-0.2in}
\end{table}

Statistics on the eventual SemLM vocabularies and semantic sequences are shown in Table~\ref{tab:vocab}. We also compare frame-chain and entity-centered SemLMs to the usual syntactic language model setting. The statistics in Table~\ref{tab:vocab} shows that they are comparable both in vocabulary size and in the total number of tokens for training. Moreover, entity-centered SemLMs have shorter sequences then frame-chain SemLMs. We also provide several examples of high-frequency augmented compound semantic frames in our generated SemLM vocabularies. All are very intuitive:

\vspace{0.05in}
\parbox{2.9in}{ \textit{want.01-know.01, \ \ \ \ \ \ agree.01-pay.01, \\ try.01-get.01, \ \ \ \ \ \ \ \ \ \ \ \ \ decline.02-comment.01, \\ wait.01-see.01, \ \ \ \ \ \ \ \ \ \ make.02-feel.01, \\ want.01(not)-give.08(up)}
               }
\subsection{Language Model Training}

\noindent {\bf NG}  We implement the N-gram model using the SRILM toolkit~\cite{stolcke2002srilm}. We also employ the well-known Kneser–Ney Smoothing~\cite{kneser1995improved} technique.

\noindent {\bf SG \& CBOW} We utilize the word2vec package to implement both SG and CBOW. In practice, we set the context window size to be 10 for SG while set the number as 5 for CBOW (both are usual settings for syntactic language models). We generate 300-dimension embeddings for both models.

\noindent {\bf LB} We use the OxLM toolkit~\cite{paul2014oxlm} with Noise-Constrastive Estimation~\cite{gutmann2010noise} for the LB model. We set the context window size to 5 and produce 150-dimension embeddings. 

\section{Evaluation} \label{sec:exp}
In this section, we first evaluate the quality of SemLMs through perplexity and a narrative cloze test. More importantly, we show that the proposed SemLMs can help improve the performance of co-reference resolution and shallow discourse parsing. 
This further proves that we successfully capture semantic sequence information which can potentially benefit a wide range of semantic related NLP tasks.

We have designed two models for SemLM: \textit{frame-chain} (\textbf{FC}) and \textit{entity-centered} (\textbf{EC}). By training on both types of sequences respectively, we implement four different language models: \textbf{TRI, SG, CBOW, LB}. We focus the evaluation efforts on these eight SemLMs.

\subsection{Quality Evaluation of SemLMs}
\noindent {\bf Datasets} We use three datasets. We first randomly sample 10\% of the New York Times Corpus documents (roughly two years of data), denoted the \textit{NYT Hold-out Data}. All our SemLMs are trained on the remaining NYT data and tested on this hold-out data. We generate semantic sequences for the training and test data using the methodology described in Sec.~\ref{sec:preprocess}.

We use PropBank data with gold frame annotations as another test set. In this case, we only generate frame-chain SemLM sequences by applying semantic unit generation techniques on gold frames, as described in Sec 5.2. When we test on \textit{Gold PropBank Data with Frame Chains}, we use frame-chain SemLMs trained from all NYT data.

Similarly, we use Ontonotes data~\cite{HMPRW06} with gold frame and co-reference annotations as the third test set, \textit{Gold Ontonotes Data with Coref Chains}. We only generate entity-centered SemLMs by applying semantic unit generation techniques on gold frames and gold co-reference chains, as described in Sec 5.2.

\noindent {\bf Baselines} We use Uni-gram (\textbf{UNI}) and Bi-gram (\textbf{BG}) as two language model baselines. In addition, we use the point-wise mutual information (PMI) for token prediction. Essentially, PMI scores each pair of tokens according to their co-occurrences. It predicts a token in the sequence by choosing the one with the highest total PMI with all other tokens in the sequence. We use the ordered PMI (\textbf{OP}) as our baseline, which is a variation of PMI by considering asymmetric counting~\cite{jans2012skip}.

\subsubsection{Perplexity}
As SemLMs are language models, it is natural to evaluate the perplexity, which is a measurement of how well a language model can predict sequences. 

Results for SemLM perplexities are presented in Table~\ref{tab:perplexity}. They are computed without considering end token (EOS). We apply tri-gram Kneser-Ney Smoothing to CBOW, SG and LB. LB consistently shows the lowest perplexities for both frame-chain and entity-centered SemLMs across all test sets. Similar to syntactic language models, perplexities are fast decreasing from UNI, BI to TRI. Also, CBOW and SG have very close perplexity results which indicate that their language modeling abilities are at the same level.

We can compare the results of our frame-chain SemLM on \textit{NYT Hold-out Data} and \textit{Gold PropBank Data with Frame Chains}, and our entity-centered SemLM on \textit{NYT Hold-out Data} and \textit{Gold Ontonotes Data with Coref Chains}. While we see differences in the results, the gap is narrow and the relative ranking of different SemLMs does not change. This indicates that the automatic SRL and Co-reference annotations added some noise but, more importantly, that the resulting SemLMs are robust to this noise as we still retain the language modeling ability for all methods.

Additionally, our ablation study removes the ``FrameNet Mapping'' step in Sec. 5.2 (``FC-FM'' and ``EC-FM'' rows), resulting in only using  PropBank frames in the vocabulary. The increase in perplexities shows that ``FrameNet Mapping'' does produce a higher level of abstraction, which is useful for language modeling.

\begin{table}
\captionsetup{belowskip=0pt,aboveskip=0pt}
\footnotesize
\caption{\textbf{Perplexities for SemLMs.} UNI, BG, TRI, CBOW, SG, LB are different language model implementations while ``FC'' and ``EC'' stand for the two SemLM models studied, respectively. ``FC-FM'' and ``EC-FM'' indicate that we removed the ``FrameNet Mapping'' step (Sec. 5.2). LB consistently produces the lowest perplexities for both frame-chain and entity-centered SemLMs.}
\label{tab:perplexity}
\centering
{
\begin{tabular}{l|l|l}
\toprule
             & \multicolumn{1}{c|}{Baselines} & \multicolumn{1}{c}{SemLMs}  \\
             & UNI \;\;\;\; BG & TRI \;\;\;\; CBOW \;\; SG \;\;\; LB  \\ \hline
\multicolumn{3}{l}{NYT Hold-out Data}                        \\
FC        & 952.1 \;\; 178.3 & 119.2 \;\; 115.4 \;\; 114.1 \;\; \textbf{108.5} \\ 
EC        & 914.7 \;\; 154.4 & 114.9 \;\; 111.8 \;\; 113.8 \;\; \textbf{109.7} \\ \hline    
\multicolumn{3}{l}{Gold PropBank Data with Frame Chains}   \\ 
FC-FM     & 992.9 \;\; 213.7 & 139.1 \;\; 135.6 \;\; 128.4 \;\; 121.8 \\ 
FC        & 970.0 \;\; 191.2 & 132.7 \;\; 126.4 \;\; 123.5 \;\; \textbf{115.4} \\ \hline
\multicolumn{3}{l}{Gold Ontonotes Data with Coref Chains}  \\ 
EC-FM     & 956.4 \;\; 187.7 & 121.1 \;\; 115.6 \;\; 117.2 \;\; 113.7 \\ 
EC        & 923.8 \;\; 163.2 & 120.5 \;\; 113.7 \;\; 115.0 \;\; \textbf{109.3} \\ \bottomrule
\end{tabular}
}
\vspace{-0.2in}
\end{table}

\begin{table*}[ht]
\captionsetup{belowskip=0pt,aboveskip=0pt}
\caption{\textbf{Narrative cloze test results for SemLMs.} UNI, BG, TRI, CBOW, SG, LB are different language model implementations while ``FC'' and ``EC'' stand for our two SemLM models, respectively. ``FC-FM'' and ``EC-FM'' mean that we remove the FrameNet mappings. ``w/o DIS'' indicates the removal of discourse makers in SemLMs. ``Rel-Impr'' indicates the relative improvement of the best performing SemLM over the strongest baseline. We evaluate on two metrics: mean reciprocal rank (MRR)/recall at 30 (Recall@30). LB outperforms other methods for both frame-chain and entity-centered SemLMs. }
\label{tab:cloze}
\centering
{
\begin{tabular}{l|ccc|cccc|c}
\toprule
\multirow{2}{*}{}  & \multicolumn{3}{c|}{Baselines} & \multicolumn{4}{c|}{SemLMs} & \multirow{2}{*}{Rel-Impr}   \\ 
         & OP    & UNI   & BG    & TRI   & CBOW  & SG  &  LB                               \\ \hline
\multicolumn{9}{c}{MRR}  \\      
\multicolumn{9}{l}{NYT Hold-out Data}   \\
FC    & 0.121 & 0.236 & 0.225 & 0.249 & 0.242 & 0.247 & \textbf{0.276} & 8.5\% \\
EC    & 0.126 & 0.235 & 0.210 & 0.242 & 0.249 & 0.249 & \textbf{0.261} & 5.9\% \\ \hline
EC w/o DIS & 0.092 & 0.191 & 0.188 & 0.212 & 0.215 & 0.216 & \textbf{0.227} & 18.8\% \\
\newcite{rudinger2015script}$^*$ & 0.083 & 0.186 & 0.181 & ----- & ----- & ----- & \textbf{0.223} & 19.9\% \\ \hline
\multicolumn{9}{l}{Gold PropBank Data with Frame Chains}   \\
FC    & 0.106 & 0.215 & 0.212 & 0.232 & 0.228 & 0.229 & \textbf{0.254} & 18.1\% \\  
FC-FM & 0.098 & 0.201 & 0.204 & 0.223 & 0.218 & 0.220 & 0.243 & --------- \\ \hline
\multicolumn{9}{l}{Gold Ontonotes Data with Coref Chains} \\
EC    & 0.122 & 0.228 & 0.213 & 0.239 & 0.247 & 0.246 & \textbf{0.257} & 12.7\% \\
EC-FM & 0.109 & 0.215 & 0.208 & 0.230 & 0.237 & 0.239 & 0.254 & --------- \\  \hline
\multicolumn{9}{c}{Recall@30}  \\ 
\multicolumn{9}{l}{NYT Hold-out Data} \\
FC    & 33.2 & 46.8 & 45.3 & 47.3 & 46.6 & 47.5 & \textbf{55.4} & 18.4\% \\
EC    & 29.4 & 43.7 & 41.6 & 44.8 & 46.5 & 46.6 & \textbf{52.0} & 19.0\% \\ \hline
\multicolumn{9}{l}{Gold PropBank Data with Frame Chains}    \\
FC    & 26.3 & 39.5 & 38.1 & 45.5 & 43.6 & 43.8 & \textbf{53.9} & 36.5\% \\ 
FC-FM & 24.4 & 37.3 & 37.3 & 42.8 & 41.9 & 42.1 & 48.2 & --------- \\ \hline 
\multicolumn{9}{l}{Gold Ontonotes Data with Coref Chains} \\
EC    & 30.6 & 42.1 & 39.7 & 46.4 & 48.3 & 48.1 & \textbf{51.5} & 22.3\% \\
EC-FM & 26.6 & 39.9 & 37.6 & 45.4 & 46.7 & 46.2 & 49.8 & --------- \\
\bottomrule
\end{tabular}
}
\vspace{-0.2in}
\end{table*}

\subsubsection{Narrative Cloze Test}
We follow the Narrative Cloze Test idea used in script learning~\cite{chambers2008unsupervised,chambers2009unsupervised}. As \newcite{rudinger2015script} points out, the narrative cloze test can be regarded as a language modeling evaluation. In the narrative cloze test, we randomly choose and remove one token from each semantic sequence in the test set. We then use language models to predict the missing token and evaluate the correctness. For all SemLMs, we use the conditional probabilities defined in Sec.~\ref{sec:slm} to get token predictions. We also use ordered PMI as an additional baseline. The narrative cloze test is conducted on the same test sets as the perplexity evaluation. We use mean reciprocal rank (MRR) and recall at 30 (Recall@30) to evaluate.

Results are provided in Table~\ref{tab:cloze}. Consistent with the results in the perplexity evaluation, LB outperforms other methods for both frame-chain and entity-centered SemLMs across all test sets. It is interesting to see that UNI performs better than BG in this prediction task. This finding is also reflected in the results reported in \newcite{rudinger2015script}. Though CBOW and SG have similar perplexity results, SG appears to be stronger in the narrative cloze test. With respect to the strongest baseline (UNI), LB achieves close to 20\% relative improvement for Recall@30 metric on NYT hold-out data. On gold data, the frame-chain SemLMs get a relative improvement of 36.5\% for Recall@30 while entity-centered SemLMs get 22.3\%. For MRR metric, the relative improvement is around half that of the Recall@30 metric.

In the narrative cloze test, we also carry out an ablation study to remove the ``FrameNet Mapping'' step in Sec. 5.2 (``FC-FM'' and ``EC-FM'' rows). The decrease in MRR and Recall@30 metrics further strengthens the argument that ``FrameNet Mapping'' is important for language modeling as it improves the generalization on frames.

We cannot directly compare with other related works~\cite{rudinger2015script,pichotta2016learning} because of the differences in data and evaluation metrics. \newcite{rudinger2015script} also use the NYT portion of the Gigaword corpus, but with Concrete annotations; \newcite{pichotta2016learning} use the English Wikipedia as their data, and Stanford NLP tools for pre-processing while we use the Illinois NLP tools. Consequently, the eventual chain statistics are different, which leads to different test instances.\footnote{\newcite{rudinger2015script} is similar to our entity-centered SemLM without discourse information. So, in Table 4, we make a rough comparison between them.} We counter this difficulty by reporting results on ``Gold PropBank Data'' and ``Gold Ontonotes Data''. We hope that these two gold annotation datasets can become standard test sets. \newcite{rudinger2015script} does share a common evaluation metric with us: MRR. If we ignore the data difference and make a rough comparison, we find that the absolute values of our results are better while \newcite{rudinger2015script} have higher relative improvement (``Rel-Impr'' in Table 4). This means that 1) the discourse information is very likely to help better model semantics 2) the discourse information may boost the baseline (UNI) more than it does for the LB model.

\begin{table}
\captionsetup{belowskip=0pt,aboveskip=0pt}
\caption{\textbf{Co-reference resolution results with entity-centered SemLM features.} ``EC'' stands for the entity-centered SemLM. ``TRI'' is the tri-gram model while ``LB'' is the log-bilinear model. ``$p_c$'' means conditional probability features and ``$em$'' represents frame embedding features. ``w/o DIS'' indicates the ablation study by removing all discourse makers for SemLMs. We conduct the experiments by adding SemLM features into the base system. We outperform the state-of-art system~\cite{wiseman2015learning}, which reports the best results on CoNLL12 dataset. The improvement achieved by ``EC\_LB ($p_c+em$)'' over the base system is statistically significant.}
\label{tab:coref}
\centering
{
\begin{tabular}{lcc}
\toprule
                               & ACE04   & CoNLL12  \\ \hline
\newcite{wiseman2015learning}  & -----   & 63.39    \\
Base~\cite{PengChRo15}         & 71.20   & 63.03    \\  
Base+EC-TRI ($p_c$)            & 71.31   & 63.14    \\
Base+EC-TRI w/o DIS            & 71.08   & 62.99    \\
Base+EC-LB \ ($p_c$)           & 71.71   & 63.42    \\
Base+EC-LB \ ($p_c+em$)        & \textbf{71.79} & \textbf{63.46}  \\
Base+EC-LB w/o DIS             & 71.12   & 63.00    \\
\bottomrule
\end{tabular}
}
\vspace{-0.2in}
\end{table}

\begin{table*}
\captionsetup{belowskip=0pt,aboveskip=0pt}
\caption{\textbf{Shallow discourse parsing results with frame-chain SemLM features.} ``FC'' stands for the frame-chain SemLM. ``TRI'' is the tri-gram model while ``LB'' is the log-bilinear model. ``$p_c$'', ``$em$'' are  conditional probability and frame embedding features, resp. ``w/o DIS'' indicates the case where we remove all discourse makers for SemLMs. We do the experiments by adding SemLM features to the base system. The improvement achieved by ``FC-LB ($p_c+em$)'' over the baseline is statistically significant.}
\label{tab:discourse}
\centering
{
\begin{tabular}{l|c|c|c|c|c|c}
\toprule
                       & \multicolumn{3}{c|}{CoNLL16 Test}      & \multicolumn{3}{c}{CoNLL16 Blind}       \\
                       & Explicit & Implicit & Overall & Explicit & Implicit & Overall   \\ \hline
Base~\cite{SPKSR15} & 89.8     & 35.6     & 60.4    & 75.8     & 31.9     & 52.3      \\
Base + FC-TRI ($q_c$)     & 90.3     & 35.8     & 60.7    & 76.4     & 32.5     & 52.9      \\
Base + FC-TRI w/o DIS      & 89.2   & 35.3 & 60.0 & 75.5 & 31.6 & 52.0   \\
Base + FC-LB \ ($q_c$)      & 90.9     & 36.2     & 61.3    & 76.8     & 32.9     & 53.4      \\
Base + FC-LB \ ($q_c+em$) & \textbf{91.1}     & \textbf{36.3}     & \textbf{61.4}    & \textbf{77.3}     & \textbf{33.2}     & \textbf{53.8}      \\
Base + FC-LB w/o DIS    & 90.1   & 35.7 & 60.6 & 76.9 & 33.0 & 53.5   \\
\bottomrule
\end{tabular}
}
\vspace{-0.2in}
\end{table*}

\subsection{Evaluation of SemLM Applications}

\subsubsection{Co-reference Resolution}
Co-reference resolution is the task of identifying mentions that refer to the same entity. To help improve its performance, we incorporate SemLM information as features into an existing co-reference resolution system. We choose the state-of-art Illinois Co-reference Resolution system~\cite{PengChRo15} as our base system. It employs a supervised joint mention detection and co-reference framework. We add additional features into the mention-pair feature set. 

Given a pair of mentions $(m_1,m_2)$ where $m_1$ appears before $m_2$, we first extract the corresponding semantic frame and the argument role label of each mention. We do this by following the procedures in Sec.~\ref{sec:preprocess}. Thus, we can get a pair of semantic frames with argument information $(\text{fa}_1,\text{fa}_2)$. We may also get an additional discourse marker between these two frames, e.g. $(\text{fa}_1,\text{dis},\text{fa}_2)$. Now, we add the following conditional probability as the feature from SemLMs:
\[ p_c = p(\text{fa}_2 | \text{fa}_1,\text{dis}). \]
We also add $p_c^2, \sqrt{p_c}$ and $1/p_c$ as features. To get the value of $p_c$, we follow the definitions in Sec.~\ref{sec:slm}, and we only use the entity-centered SemLM here as its vocabulary covers frames with argument labels. For the neural language model implementations (CBOW, SG and LB), we also include frame embeddings as additional features.

We evaluate the effect of the added SemLM features on two co-reference benchmark datasets: ACE04~\cite{nist2-ace} and CoNLL12~\cite{PMXUZ12}. We use the standard split of 268 training documents, 68 development documents, and 106 testing documents for ACE04 data~\cite{CWHM07,BengtsonRo08}. For CoNLL12 data, we follow the train and test document split from CoNLL-2012 Shared Task. We report CoNLL AVG for results (average of MUC, B$^3$, and CEAF$_e$ metrics), using the v7.0 scorer provided by the CoNLL-2012 Shared Task.

Co-reference resolution results with entity-centered SemLM features are shown in Table~\ref{tab:coref}. Tri-grams with conditional probability features improve the performance by a small margin, while the log-bilinear model achieves a 0.4-0.5 F1 points improvement. By employing log-bilinear model embeddings, we further improve the numbers and  we outperform the best reported results on the CoNLL12 dataset~\cite{wiseman2015learning}.

In addition, we carry out ablation studies to remove all discourse makers during the language modeling process. We re-train our models and study their effects on the generated features. Table~\ref{tab:coref} (``w/o DIS" rows) shows that without discourse information, the SemLM features would hurt the overall performance, thus proving the necessity of considering discourse for semantic language models.

\subsubsection{Shallow Discourse Parsing}
Shallow discourse parsing is the task of identifying explicit and implicit discourse connectives, determine their senses and their discourse arguments. In order to show that SemLM can help improve shallow discourse parsing, we evaluate on identifying the correct sense of discourse connectives (both explicit and implicit ones).

We choose~\newcite{SPKSR15}, which uses a supervised pipeline approach, as our base system. The system extracts context features for potential discourse connectives and applies the discourse connective sense classifier. Consider an explicit connective ``dis''; we extract the semantic frames that are closest to it (left and right), resulting in the sequence [f$_1$, dis, f$_2$] by following the procedures described in Sec.~\ref{sec:preprocess}. We then add the following conditional probabilities as features.  Compute
\[ q_c = p(\text{dis} | \text{f}_1, \text{f}_2). \]
and, similar to what we do for co-reference resolution, we add $q_c, q_c^2, \sqrt{q_c}, 1/q_c$ as conditional probability features, which can be computed following the definitions in Sec.~\ref{sec:slm}.  We also include frame embeddings as additional features. We only use frame-chain SemLMs here. 

We evaluate on CoNLL16~\cite{xue2015conll} test and blind sets, following the train and development document split from the Shared Task, and report F1 using the official shared task scorer.

Table~\ref{tab:discourse} shows the results for shallow discourse parsing with SemLM features. Tri-gram with conditional probability features improve the performance for both explicit and implicit connective sense classifiers. Log-bilinear model with conditional probability features achieves even better results, and frame embeddings further improve the numbers. SemLMs improve relatively more on explicit connectives than on implicit ones.

We also show an ablation study in the same setting as we did for co-reference, i.e. removing discourse information (``w/o DIS" rows). While our LB model can still exhibit  improvement over the base system, its performance is lower than the proposed discourse driven version, which means that discourse information improves the expressiveness of semantic language models.

\section{Conclusion} \label{sec:conclusion}
The paper builds two types of discourse driven semantic language models with four different language model implementations that make use of neural embeddings for semantic frames. We use perplexity and a narrative cloze test to prove that the proposed SemLMs have a good level of abstraction and are of high quality, and then apply them successfully to the two challenging  tasks of co-reference resolution and shallow discourse parsing, exhibiting improvements over state-of-the-art systems. In future work, we plan to apply SemLMs to other semantic related NLP tasks e.g. machine translation and question answering.

\section*{Acknowledgments}
The authors would like to thank Christos Christodoulopoulos and Eric Horn for comments that helped to improve this work.
This work is supported by Contract HR0011-15-2-0025 with the US Defense Advanced Research Projects Agency (DARPA). Approved for Public Release, Distribution Unlimited. The views expressed are those of the authors and do not reflect the official policy or position of the Department of Defense or the U.S. Government. This material is also based upon work supported by the U.S. Department of Homeland Security under Award Number 2009-ST-061-CCI002-07.

\bibliographystyle{acl2016}
\bibliography{ccg-compact,cited-compact,new}

\end{document}